\pdfoutput=1

\documentclass[11pt]{article}

\usepackage{acl}

\usepackage{times}
\usepackage{latexsym}
\usepackage{graphicx}
\usepackage{booktabs}
\usepackage{xcolor}
\usepackage{amsmath}
\usepackage{footmisc}
\usepackage{mathtools}
\usepackage{stfloats}

\definecolor{factual}{HTML}{4878D0}
\definecolor{factuallight}{HTML}{7399E6}
\definecolor{sna}{HTML}{EE854a}
\definecolor{wordplay}{HTML}{77E6B9}
\definecolor{wordmeaning}{HTML}{d65f5f}
\definecolor{fillblank}{HTML}{956cb4}
\definecolor{historical}{HTML}{8c613c}
\definecolor{historicallight}{HTML}{DBA670}
\definecolor{dependent}{HTML}{dc7ec0}
\definecolor{abbr}{HTML}{797979}
\definecolor{abbrlight}{HTML}{C2C2C2}
\definecolor{prefixsuffix}{HTML}{d5bb67}
\definecolor{crosslingual}{HTML}{4878D0}
\definecolor{crosslinguallight}{HTML}{90DBFA}

\usepackage[T1]{fontenc}

\usepackage[utf8]{inputenc}

\usepackage{microtype}

\usepackage{enumitem}
\setitemize{noitemsep,topsep=2pt,parsep=0pt,partopsep=0pt}
\setenumerate{noitemsep,topsep=2pt,parsep=0pt,partopsep=0pt}

\title{Down and Across: Introducing Crossword-Solving as a New NLP Benchmark}

\author{
Saurabh Kulshreshtha, Olga Kovaleva, Namrata Shivagunde, and Anna Rumshisky\\
Department of Computer Science\\
University of Massachusetts Lowell\\
\texttt{\{skul,okovalev,nshivagu,arum\}@cs.uml.edu}\\
}

\begin{document}
\maketitle
\begin{abstract}
   Solving crossword puzzles requires diverse reasoning capabilities, access to a vast amount of knowledge about language and the world, and the ability to satisfy the constraints imposed by the structure of the puzzle. In this work, we introduce solving crossword puzzles as a new natural language understanding task. We release the specification of a corpus of crossword puzzles collected from the New York Times daily crossword spanning 25 years and comprised of a total of around nine thousand puzzles. These puzzles include a diverse set of clues: historic, factual, word meaning, synonyms/antonyms, fill-in-the-blank, abbreviations, prefixes/suffixes, wordplay, and cross-lingual, as well as clues that depend on the answers to other clues. We separately release the clue-answer pairs from these puzzles as an open-domain question answering dataset containing over half a million unique clue-answer pairs. For the question answering task, our baselines include several sequence-to-sequence and retrieval-based generative models. We also introduce a non-parametric constraint satisfaction baseline for solving the entire crossword puzzle. Finally, we propose an evaluation framework which consists of several complementary performance metrics.
\end{abstract}

\section{Introduction}
Recent breakthroughs in NLP established high standards for the performance of machine learning methods across a variety of tasks. However, even state-of-the-art models demonstrate fragility \cite{wallace2019universal} and exhibit sensitivity to shallow data patterns 
\cite{mccoy2019right,zellers2019hellaswag,jin2020bert,si2019does,sugawara2020assessing,yogatama2019learning,niven2019probing}.
This has led to a growing demand for successively more challenging tasks. 

One of the important tasks in natural language understanding is question answering (QA), with many recent datasets created to address different different aspects of this task \cite{yang2018hotpotqa, rajpurkar2016squad, kwiatkowski2019natural, zellers2019hellaswag, dua2019orb, rogers2021taxonomy}. 
%
%
There are two main forms of question answering (QA): extractive QA and open-domain QA. In extractive QA, a passage that answers the question is provided as input to the system along with the question.  In open-domain QA, only the question is provided as input, and the answer must be generated either through memorized knowledge or via some form of explicit information retrieval over a large text collection which may contain answers. 

The task of answering clues in a crossword is a form of open-domain question answering.
Once a human or an open-domain QA system generates a few possible answer candidates for each clue, one of these candidates may form the correct answer to a word slot in the crossword grid, if the candidate meets the constraints of the crossword grid.


Solving a crossword puzzle is therefore a challenging task which requires (1) finding answers to a variety of clues that require extensive language and world knowledge, and (2) the ability to produce answer strings that meet the constraints 
of the crossword grid, including length of word slots and character overlap with other answers in the puzzle.

Our contributions in this work are as follows:
\begin{itemize}
\item We introduce a new natural language understanding task of solving crossword puzzles, along with the specification of a dataset of New York Times crosswords from Dec. 1, 1993 to Dec. 31, 2018.
\item We propose an evaluation framework which consists of several complementary performance metrics.
\item We release the collection of clue-answer pairs as a new open-domain QA dataset.
\item We provide baselines for the proposed crossword task and the new QA task, including several sequence-to-sequence and retrieval-augmented generative Transformer models, with a constraint satisfaction crossword solver.
\end{itemize}


\begin{figure*}[ht]
    \centering
    \includegraphics[width=0.7\linewidth]{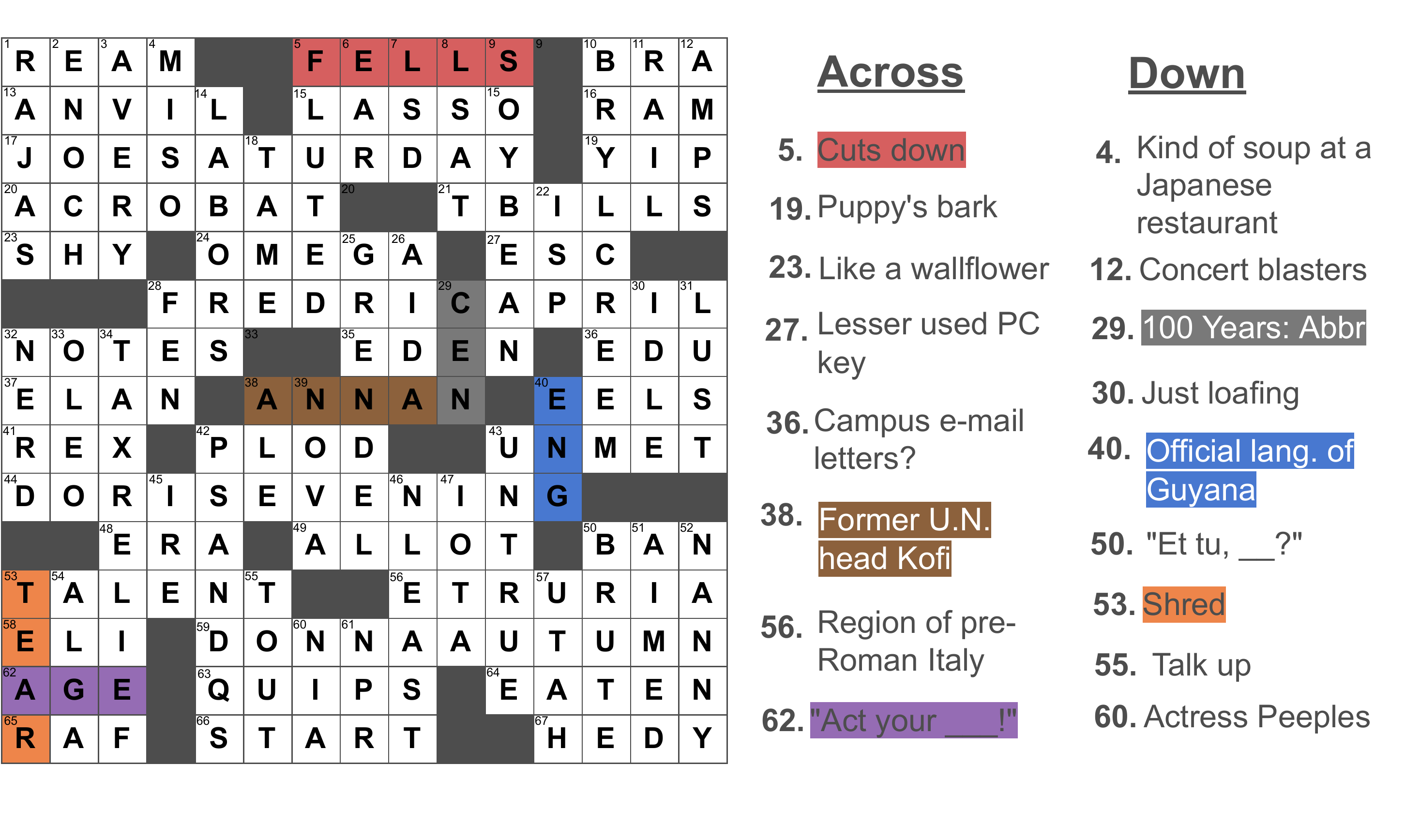}
\vspace{-0.7cm}
    \caption{Crossword puzzle example. A few clues from the puzzle have been provided on the right, they are filled horizontally (Across) or vertically (Down) in the crossword grid. The clue number tells the player where in the grid the answer needs to be filled in. Some of these clue and their answers have further been highlighted with different colors which belong to different clue categories as described in Section \ref{sec:clue_types}, color-coded in accordance with  \autoref{fig:piechart_1000annotation}. Highlight colors denote distinct clue categories: red for word meaning clues, purple for fill-in-the blank clue, orange for synonym/antonym, blue for factoid question type, grey for abbreviation and brown for historical. Source: New York Times daily crossword which appeared on the July 7, 2009. Copyright of The New York Times, 2009.}
    \label{fig:crossword_example}.
\end{figure*}

\section{Related Work}



Our work is in line with open-domain QA benchmarks.
Examples of such tasks include datasets where each question can be answered using information contained in a relevant Wikipedia article \cite{yang2015wikiqa,kwiatkowski2019natural,yang2018hotpotqa}. 
Several QA tasks have been designed to require multi-hop reasoning over structured knowledge bases \cite{berant2013semantic,bordes2015large}. The main limitation of such datasets is that their question types are mostly factual.
%
%
Crossword clues differ from these efforts in that they combine a variety of different reasoning types.

Another line of research that is relevant to our work explores the problem of solving Sudoku puzzles since it is also a constraint satisfaction problem.
Most sudoku puzzles can be efficiently solved by algorithms that take advantage of the fixed input size and do not rely on machine learning methods \cite{simonis2005sudoku}. The machine learning attempts for solving Sudoku puzzles have been inspired by convolutional \cite{mehta2021reinforcement} and recurrent relational networks \cite{palm2017recurrent}.
Unlike Sudoku, however, where the grids have the same structure, shape and constraints,
crossword puzzles have arbitrary shape and internal structure and rely on answers to natural language questions that require reasoning over different kinds of world knowledge.

Several previous studies have treated crossword puzzle solving as a constraint satisfaction problem (CSP) \cite{LITTMAN200223,ernandes2005,ginsberg2011dr}. \citet{LITTMAN200223}'s Proverb system incorporates a variety of information retrieval modules to generate candidate answers. The Database module searches a large database of historical clue-answer pairs to retrieve the answer candidates. They find very poor crossword-solving performance in ablation experiments where they limit their answer candidate generator modules to not use historical clue-answer databases. WebCrow \cite{ernandes2005} builds upon Proverb and makes improvements to the database retriever module augmented with a new web module which searches the web for snippets that may contain answers. It allows partial matching to retrieve clues-answer pairs in the historical database that do not perfectly overlap with the query clue. 
Dr. Fill system proposed by \citet{ginsberg2011dr} treats each crossword puzzle as a singly-weighted CSP. Similarly to prior work, Dr. Fill relies on a large set of historical clue-answer pairs (up to 5M) collected over multiple years from the past puzzles by applying direct lookup and a variety of heuristics. One common design aspect of all these solvers is to generate answer candidates independently from the crossword structure and later use a separate puzzle solver to fill in the actual grid. In our work, we  partition the task of crossword solving similarly.

\citet{barlacchi14} and \citet{severyn15} observe that the most important source of candidate answers for a given clue is a large database of historical clue-answer pairs and introduce methods to better search these databases. \citet{barlacchi14} apply a BM25 retrieval model to generate clue lists similar to the query clue from historical  clue-answer database, where the generated clues get further refined through application of re-ranking models.
\citet{severyn15} introduce a distributional neural network to compute similarities between clues trained over a large scale dataset of clues that they introduce.

In contrast to the previous work, our goal in this work is to motivate solver systems to generate answers organically, just like a human might, rather than obtain answers via the lookup in historical clue-answer databases. The answers could be generated either from memory of having read something relevant, using world knowledge and language understanding, or by searching encyclopedic sources such as Wikipedia or a dictionary with relevant queries.

\section{Task and Dataset}

For the purposes of our task, crosswords are defined as 
word puzzles with a given rectangular grid of white- and black-shaded squares. The goal is to fill the white squares with letters, forming words or phrases by solving textual clues which lead to the answers. The answer words and phrases are placed in the grid from left to right ("Across") and from top to bottom ("Down"). The shaded squares are used to separate the words or phrases. Usually, the white spaces and punctuation are removed from the answer phrases.
A sample crossword puzzle is given in \autoref{fig:crossword_example}. Note that the answers can include named entities and abbreviations, and at times require the exact grammatical form, such as the correct verb tense or the plural noun.

Solving a crossword puzzle is a complex task that requires generating the right answer candidates and selecting those that satisfy the puzzle constraints. Similar to prior work, we divide the task of solving a crossword puzzle into two subtasks, to be evaluated separately. The first subtask can be viewed as a question answering task, where a system is trained to generate a set of candidate answers for a given clue without taking into account any interdependencies between answers. The second subtask involves solving the entire crossword puzzle, i.e., filling out the crossword grid with a subset of candidate answers generated in the previous step.

The two tasks could be solved separately or in an end-to-end fashion.
In contrast to prior work \cite{ernandes2005,ginsberg2011dr}, our clue-answer data is linked directly with our puzzle-solving data, so no data leakage is possible between the QA training data and the crossword-solving test data. 
%
In the present work, we propose a separate solver for each task. 
We provide details on the challenges of implementing an end-to-end solver in the discussion section.

\subsection{NYT Crossword Collection}
Our dataset is sourced from the New York Times, which has been featuring a daily crossword puzzle since 1942. We worked with daily puzzles in the date range from December 1, 1993 through December 31, 2018 inclusive. All the crossword puzzles in our corpus are available to play through the New York Times games website \footnote{\url{https://www.nytimes.com/crosswords}}. We release two separate specifications of the dataset corresponding to the subtasks described above: the NYT Crossword Puzzle dataset and the NYT Clue-Answer dataset.\footnote{Details for dataset access will be made available at \url{https://github.com/text-machine-lab/xword_benchmark}. We are currently finalizing the agreement with the New York Times to release this dataset.}

 
There are a few details that are specific to the NYT daily crossword. First, the clue and the answer must agree in tense, part of speech, and even language, so that the clue and answer could easily be substituted for each other in a sentence. Second, abbreviated clues indicate abbreviated answers. Further, clues that end in a question mark indicate a play on words in the clue or the answer. There are also a lot of short words that appear in crosswords much more often than in real life. These 3- and 4-letter words, referred to as crosswordese, can be very helpful in solving the puzzles. Finally, every Sunday through Thursday NYT crossword puzzle has a theme, something that unites the puzzle’s longest answers. Theme answers are always found in symmetrical places in the grid.

\paragraph{Crossword Puzzle Dataset.} The dataset consists of 9152 puzzles, split into the training, validation, and test subsets in the 80/10/10 ratio which give us 7293/922/941 puzzles in each set. 
We removed the total of 50/61 special puzzles from the validation and test splits, respectively, because they used non-standard rules for filling in the answers, such as L-shaped word slots or allowing cells to be filled with multiple characters (called rebus entries).




Most NYT crossword grids have a square shape of $15{\times}15$ cells, with the exception of Sunday-released crosswords being $21{\times}21$ cells. Other shapes combined account for less than $3\%$ of the data. The vast majority of both clues and answers are short, with over 76\% of clues consisting of a single word. For traditional sequence-to-sequence modeling such conciseness imposes an additional challenge, as there is very little context provided to the model. In most puzzles, over 80\% of the grid cells are filled and \textit{every} character is an intersection of two answers. Such high answer inter-dependency suggests a high cost of answer misprediction, as errors affect a larger number of intersecting words. More detailed statistics on the dataset are given in \autoref{tab:stats}.
\paragraph{Clue-Answer Dataset.} We generate an open-domain question answering dataset consisting solely of clue-answer pairs from the respective splits of the Crossword Puzzle dataset described above (including the special puzzles). Within each of the splits, we only keep unique clue-answer pairs and remove all duplicates. 
However, certain clues may still be shared between the puzzles contained in different splits. We therefore remove from the training data the clue-answer pairs which are found in the test or validation data. 
This ensures that the model can not trivially recall the answers to the overlapping clues while predicting for the test and validation splits. 

This produces the total of $578$k clue-answer pairs, with $433$k/$72$k/$72$k examples in the train/validation/test splits, respectively. Since certain answers consist of phrases and multiple words that are merged into a single string (such as "VERYFAST"), we further postprocess the answers by splitting the strings into individual words using a dictionary. Out of all the possible word splits of a given string we pick the one that has the smallest number of words. If there are multiple solutions, we select the split with the highest average word frequency. Examples of a variety of clues found in this dataset are given in the following section.

\subsection{Clue types} \label{sec:clue_types}

To provide more insight into the diversity of the clue types and the complexity of the task, we categorize all the clues into multiple classes, which we describe below.

\paragraph{Factual.} Clues that encode encyclopedic knowledge and typically can be answered using resources such as Wikipedia (e.g. \textit{Clue: South Carolina State tree, Answer: PALMETTO}). This type of clue is the closest to the questions found in open-domain QA datasets. Note that the facts required to solve some of the clues implicitly depend on the date when a given crossword was released. For instance, the clue "\textit{President of Brazil}" has a time-dependent answer.

\paragraph{Historical.} Clues that require the knowledge of historical facts and temporal relations between events. (e.g. \textit{Clue: Automobile pioneer, Answer: BENZ}).

\paragraph{Word meaning.} Clues that exploit general vocabulary knowledge and can typically be resolved using a dictionary. (e.g. \textit{Clue: Opposing sides, Answer: FOES}).

\paragraph{Synonyms/Antonyms.} Clues that focus on paraphrasing and synonymy relations (e.g. \textit{Clue: Prognosticators, Answer: SEERS}). In most cases, such clues can be solved with a thesaurus.

\paragraph{Fill in the blank.} Clues formulated as a cloze task (e.g. \textit{Clue: Magna Cum \_\_, Answer: LAUDE}). Fill-in-the-blank clues are expected to be easy to solve for the models trained with the masked language modeling objective \cite{devlin2019bert}.

\paragraph{Abbreviations.} Clues answered with acronyms (e.g. \textit{Clue: (Abbr.) Old Communist state, Answer: USSR}). Abbreviation clues are marked with "\textit{Abbr.}" label.

\paragraph{Prefix/Suffix.}Clues that suggest the answer is a suffix or prefix. (e.g. \textit{Clue: Suffix with mountain, Answer: EER})

\paragraph{Wordplay.} Clues that rely on wordplay, anagrams, or puns /  pronunciation similarities (e.g. \textit{Clue: Consider an imaginary animal, Answer: BEAR IN MIND}). In a lot of cases, wordplay clues involve jokes and exploit different possible meanings and contexts for the same word.

\paragraph{Cross-lingual.} Clues that either explicitly use words from other languages, or imply a specific language-dependent form of the answer.
(e.g. \textit{Clue: Sunrise dirección, Answer: ESTE}).

\paragraph{Clues dependent on other clues.} Clues the answer to which can be provided only after a different clue has been solved (e.g. \textit{Clue: Last words of 45 Across}). Although rare, this category of clues suggests that the entire puzzle has to be solved in certain order.

\begin{figure*}[!ht]
    \centering
    \includegraphics[width=0.7\linewidth]{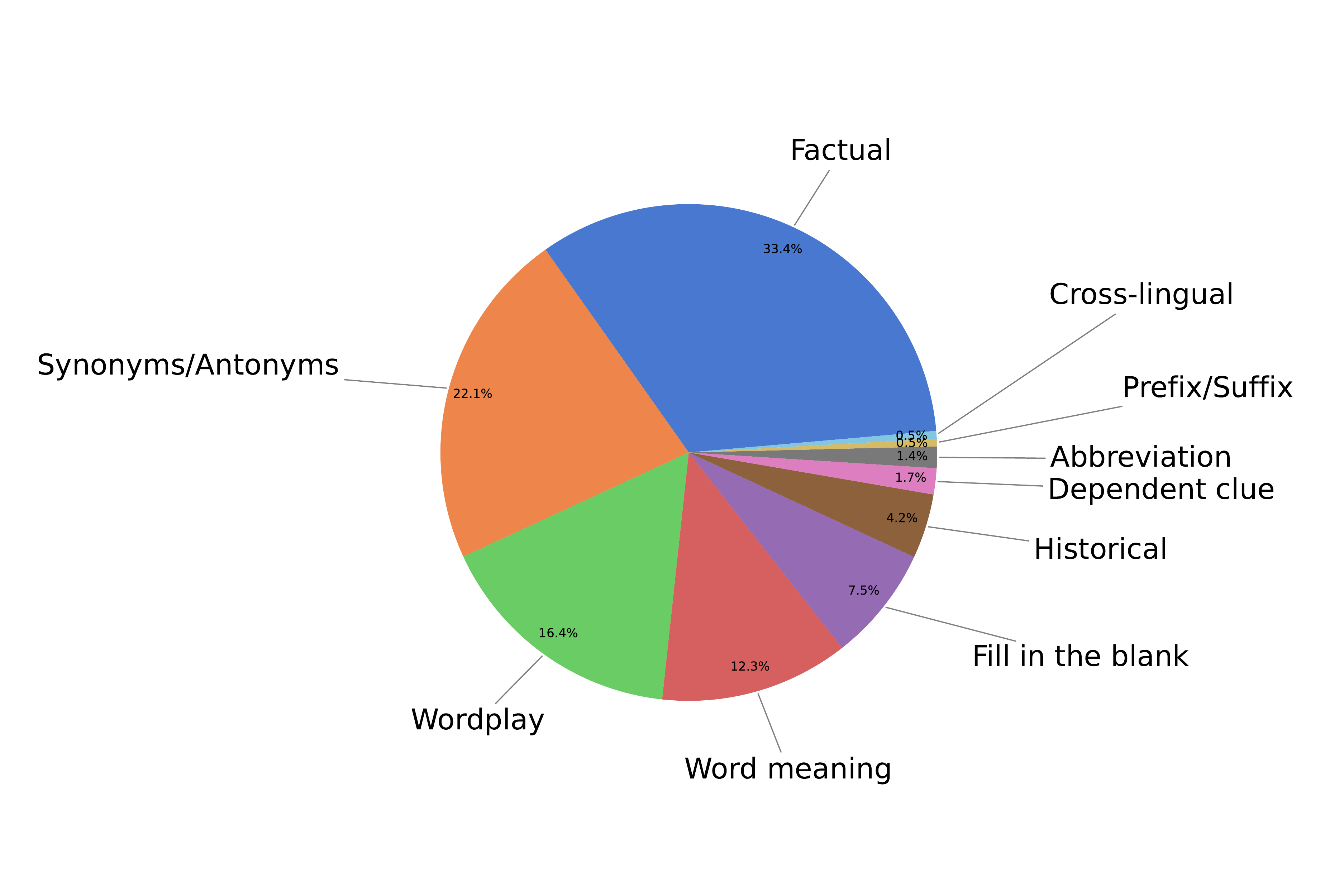}
    \vspace{-0.7cm}
    \caption{Class distribution of the 1000 manually annotated test examples.}
    \label{fig:piechart_1000annotation}
\end{figure*}


To understand the distribution of these classes, we randomly selected 1000 examples from the test split of the data and manually annotated them. \autoref{fig:piechart_1000annotation} illustrates the class distribution of the annotated examples, showing that the Factual class covers a little over a third of all examples. The synonyms/antonyms, word meaning and wordplay classes taken together comprise 50\% of the data. The remaining 20\% are taken by fill-in-the-blank and historical clues, as well as the low-frequency classes (comprising less than or around 1\%), which include abbreviation, dependent, prefix/suffix and cross-lingual clues. We illustrate each one of these classes in the \autoref{fig:crossword_example}.

\begin{table}[th!]
\footnotesize
\begin{tabular}{p{2.3cm}p{1.2cm}p{1.2cm}p{1.2cm}}
\toprule
 & \textbf{Train} & \textbf{Validation} & \textbf{Test} \\
 \midrule
 & \multicolumn{3}{l}{Clue-Answer dataset} \\
 \midrule
\# clues & 4,33,033 & 72,303 & 72,939 \\
avg/median clue length (words) & 4.0/3 & 4.2/4 & 4.2/4 \\
avg/median ans. length (chars) & 5.5/5 & 5.7/5 & 5.6/5 \\
avg/median ans. length (words) & 1.3/1 & 1.3/1 & 1.3/1 \\
\midrule
 & \multicolumn{3}{l}{Crossword Puzzle dataset} \\
 \midrule
\# puzzles & 7,293 & 872 & 879 \\
avg/median \# of clues & 83.5/76 & 83.6/76 & 82.9/76 \\
avg cols${\times}$rows & 15.9${\times}$15.9 & 15.9${\times}$15.9 & 15.8${\times}$15.8 \\
\% of cells filled & 82.20\% & 80.20\%  & 81.20\% \\
\bottomrule
\end{tabular}
\caption{The full statistics on the two versions of the released datasets.}
\label{tab:stats}
\end{table}

\subsection{Evaluation metrics} \label{sec:evaluation_metrics}

In this section, we describe the performance metrics we introduce for the two subtasks.

\paragraph{Clue-Answer Task.}
For the clue-answer task, we use the following metrics:
\begin{itemize}
\item \textbf{Exact Match (EM)}. Model output matches the ground-truth answer exactly.
\item \textbf{Contains (In)}. Model output contains the ground-truth answer as a contiguous substring
\end{itemize}

\noindent
Since the ground-truth answers do not contain diacritics, accents, punctuation and whitespace characters, we also consider normalized versions of the above metrics, in which these are stripped from the model output prior to computing the metric.  We will refer to them as
\textbf{EM\textsubscript{norm}} and \textbf{In\textsubscript{norm}},

We report these metrics for top-$k$ predictions, where $k$ varies from 1 to 20.

\paragraph{Crossword Puzzle Task.} 
To evaluate the performance of the crossword puzzle solver, we propose to compute the following two metrics:

\begin{itemize}
\item \textbf{Character Accuracy (Acc\textsubscript{char})}. Percentage of characters in the predicted crossword solution that match the ground-truth solution.
\item \textbf{Word Accuracy (Acc\textsubscript{word})}. Percentage of words in the predicted crossword solution that match the ground-truth solution.
\end{itemize}


Since the clue-answering system might not be able to generate the right answers for some of the clues, it may only be possible to produce a partial solution to a puzzle. The crossword puzzle solver will fail to produce a solution when the answer candidate list for a clue does not contain the correct answer. To prevent this from happening, the character cells which belong to that clue's answer must be removed from the puzzle grid, unless the characters are shared by other clues.  We propose two additional metrics to track what percentage of the puzzle needs to be redacted to produce a partial solution:

\begin{itemize}
\item \textbf{Word Removal (Rem\textsubscript{word})}. \% of words that need to be removed from the puzzle to produce a partial solution.

\item \textbf{Character Removal (Rem\textsubscript{word})}. \% of characters that need to be removed from the puzzle grid to produce a partial solution.
\end{itemize}

The motivation for introducing the removal metrics is to indicate the amount of constraint relaxation. For instance, a completely relaxed puzzle grid, where many character cells have been removed, such that the grid has no word intersection constraints left, could be considered "solved" by selecting any candidates from the answer candidate lists at random. However, this solution will mostly be incorrect when compared to the gold puzzle solution. As the word and character removal percentage increases, the potential for correctly solving the remaining puzzle is expected to decrease, since the  under-constrained answer cells in the grid can be incorrectly filled by other candidates (which may not be the right answers). The removal metrics are thus complementary to word and character level accuracy.


\section{Baselines}


Our baseline approach is a two-step solution that treats each subtask separately.  We first develop a set of baseline systems that solve the question answering problem, ignoring the grid-imposed answer interdependencies. 
We use seq-to-seq and retrieval-augmented Transformer baselines for this subtask.
We feed generated answer candidates to a crossword solver in order to complete the puzzle and evaluate the produced puzzle solutions.

\subsection{Clue-Answer Task Baselines}

\paragraph{Sequence-to-sequence baselines.} 
We fine-tune two sequence-to-sequence models on the clue-answer training data. We select two widely known models, BART \cite{lewis2019bart} and T5 \cite{raffel2019exploring}, which achieved state-of-the-art results on a set of generative tasks, including specifically abstractive QA involving commonsense and multi-hop reasoning \cite{fan2019eli5,khashabi2018looking,zhang2018record}.

We train both models for 8 epochs with the learning rate of $5\times10^{-5}$, and a batch size of 60. \footnote{We use BART-large with approximately 406M parameters and T5-base model with approximately 220M parameters, respectively.}

\begin{table*}[ht]
    \small
    \centering
    \begin{tabular}{lllll|llll|llll}
    \toprule
     & \multicolumn{4}{c}{Top-1} & \multicolumn{4}{c}{Top-10} & \multicolumn{4}{c}{Top-20}   \\
     & \textbf{EM} & \textbf{EM\textsubscript{norm}} & \textbf{In} & \textbf{In\textsubscript{norm}} & \textbf{EM} & \textbf{EM\textsubscript{norm}} & \textbf{In} & \textbf{In\textsubscript{norm}} & \textbf{EM} & \textbf{EM\textsubscript{norm}} & \textbf{In} & \textbf{In\textsubscript{norm}}   \\
     \midrule
    T5-base & 8.4 & 9.5 & 8.7 & 9.9 & 18.7 & 20.8 & 19.8 & 22.0 & 22.2 & 24.6 & 23.8 & 26.3   \\
    BART-large & 13.8 & 16.1 & 15.0 & 17.6 & 31.0 & 36.7 & 32.4 & 38.0 & 34.0 & 40.1 & 35.3 & 41.3  \\
    RAG wiki & 24.2 & 26.0 & 24.9 & 26.7 & 46.8 & 49.8 & 48.6 & 51.6 & 50.6 & 53.9 & 53.4 & 56.7 \\
    RAG dict & 24.0 & 25.8 & 24.6 & 26.5 & 46.0 & 48.9 & 48.0 & 50.9 & 50.0 & 53.2 & 53.0 & 56.2 \\
    \bottomrule
    \end{tabular}
    \caption{Performance of baseline systems on the Clue Answering dataset. \textit{EM} and \textit{In} stand for the ``Exact-match'' and ``Contains'' metrics as described in Section \ref{sec:evaluation_metrics}. The computed metrics are shown for top-1, top-10, and top-20 predictions for a given model. }
    \label{tab:clue_answering_all_models}
\end{table*}

\paragraph{Retrieval-augmented generation.} 
T5 and BART store world knowledge implicitly in their parameters and are known to hallucinate facts \cite{maynez-etal-2020-faithfulness}. Recently, a new method called retrieval-augmented generation (RAG) \cite{lewis2020retrieval} has been introduced for open-domain question answering. This method involves a Transformer encoder to encode the question and a decoder to generate the answer \cite{vaswani2017attention}, but the encoded query is supplemented with relevant excerpts retrieved from an external textual corpus via Maximum Inner Product Search (MIPS); the entire neural network is trained end-to-end. Due to a built-in retrieval mechanism for performing a soft search over a large collection of external documents, such systems are capable of producing stronger results on knowledge-intensive open-domain question answering tasks than the vanilla sequence-to-sequence generative models and are more factually accurate
\cite{shuster2021retrieval}.
 Motivated by this, we train RAG models to extract knowledge from two separate external sources of knowledge: 
 \begin{enumerate}
 \item[(a)] \textbf{RAG-wiki} uses a full Wikipedia dump from December 2018. Following existing work \citet{lewis2020retrieval,karpukhin-etal-2020-dense,lee-etal-2019-latent}, each Wikipedia article is split into disjoint 100-word chunks, resulting in a total of 21M passages.
 \item[(b)] \textbf{RAG-dict} uses several English dictionaries and thesauri sources, including Wiktionary\footnote{\url{https://www.wiktionary.org/}}, Merriam-Webster\footnote{\url{https://dictionaryapi.com/}}, and Google’s English dictionary by Oxford Languages.\footnote{Accessed via \url{https://dictionaryapi.dev/}.}
 \end{enumerate}
For both of these models, we use the retriever embeddings pretrained on the Natural Questions corpus \citet{nq} in order to prime the MIPS retrieval to return meaningful entries \cite{lewis2020retrieval}.  We train with a batch size of 8, label smoothing set to 0.1, dropout probability of 0.1, weight decay rate of 0.001, and a learning rate of $3\times10^{-5}$ for 8 epochs.

\subsection{Crossword Puzzle Task}
A crossword puzzle can be cast as an instance of a satisfiability problem, and its solution represents a particular character assignment so that all the constraints of the puzzle are met.
Under such formulation, three main conditions have to be satisfied: (1) the answer candidates for every clue must come from a set of words that  answer the question, (2) they must have the exact length specified by the corresponding grid entry, and (3) for every pair of words that intersect in the puzzle grid, acceptable word assignments must have the same character at the intersection offset.

This class of problems can be modelled through Satisfiability Modulo Theories (SMT).
SMT is a generalization of Boolean Satisfiability problem (SAT) in which some of the binary variables are replaced by first-order logic predicates over a set of non-binary variables. In the case of crosswords, a variable represents one character in the crossword grid which can be assigned a single letter of the English alphabet and 0 through 9 digit values. This is further subject to the constraints mentioned above which can be formulated with the equality operator and Boolean logical operators: \texttt{AND} and \texttt{OR}. For example, a word slot of length 3 where the candidate answers are "ESC", "DEL" or "CMD" can be formalised as:
\begin{gather*}
\scalebox{0.8}{$\{v_1\!=\!E\,\texttt{AND}\,v_2\!=\!S\,\texttt{AND}\,v_3\!=\!C\,\}$}\\
\scalebox{0.8}{$OR$}\\
\scalebox{0.8}{$\{v_1\!=\!D\,\texttt{AND}\,v_2\!=\!E\,\texttt{AND}\,v_3\!=\!L\,\}$}\\
\scalebox{0.8}{$OR$}\\
\scalebox{0.8}{$\{v_1\!=\!C\,\texttt{AND}\,v_2\!=\!M\,\texttt{AND}\,v_3\!=\!D\,\}$}
\end{gather*}
To solve the entire crossword puzzle, we use the formulation that treats this as an SMT problem. We modify an open source implementation\footnote{\url{https://github.com/pncnmnp/Crossword-Solver}\label{z3solver}} of this formulation based on Z3 SMT solver \cite{z3}. The answer length and intersection constraints are imposed on the variable assignment, as specified by the input crossword grid.


We take the top-$k$ predictions from our baseline models 
and for each prediction, select all possible substrings of required length as answer candidates. For simplicity, we exclude from our consideration all the crosswords with a single cell containing more than one English letter in it.

Our current baseline constraint satisfaction solver is limited in that it simply returns "not-satisfied" (\texttt{nosat}) for a puzzle where no valid solution exists, that is, when \textit{all} the hard constraints of the puzzle are not met by the inputs. Since the candidate lists for certain clues might not meet all the constraints, this results in a \texttt{nosat} solution for almost all crossword puzzles, and we are not able to extract partial solutions. To bypass this issue and produce partial solutions, we pre-filter each clue with an oracle that only allows those clues into the SMT solver for which the actual answer is available as one of the candidates.


\section{Results} \label{sec:results}
\subsection{Clue-Answer Task}
In \autoref{tab:clue_answering_all_models} we report the Top-1, Top-10 and Top-20 match accuracies for the four evaluation metrics defined in Section \ref{sec:evaluation_metrics}.

Our results (\autoref{tab:clue_answering_all_models}) suggest a high difficulty of the clue-answer dataset, with the best achieved accuracy metric staying under 30\% for the top-1 model prediction. Even top-20 predictions have an almost 40\% chance of not containing the ground-truth answer anywhere within the generated strings.
Generative Transformer models such as T5-base and BART-large perform poorly on the clue-answer task, however, the model accuracy across most metrics almost doubles when switching from T5-base (with 220M parameters) to BART-large (with 400M parameter). 

Our strongest baseline, RAG-wiki and RAG-dict, achieve 50.6 and 50.0 exact-match accuracies on the clue-answer dataset, respectively. 
The ${In}_{norm}$ score, which looks at whether any substrings in the generated answer match the ground truth -- and which can be seen an upper bound on the model's ability to solve the puzzle -- is slightly higher, at 56.7 for RAG-wiki and 56.2 for RAG-dict. 

Not surprisingly, these results show that the additional step of retrieving Wikipedia or dictionary entries increases the accuracy considerably compared to the fine-tuned sequence-to-sequence models such as BART which store this information in its parameters.
%
%
The normalized metrics which remove diacritics, punctuation and whitespace bring the accuracy up by 2-6\%, depending on the model.

We examined the top-20 exact-match predictions generated by RAG-wiki and RAG-dict and find that both models are in agreement in terms of answer matches for around 85\% of the test set. In other words, both models either correctly predict the ground truth answer or both fail to do so. 

\begin{table}[t]
    \small
    \centering
    \begin{tabular}{l|ll|ll}
        \toprule
        Model & \multicolumn{2}{c}{Solving Accuracy} & \multicolumn{2}{c}{Puzzle Removed} \\
        & \textbf{Acc\textsubscript{word}} & \textbf{Acc\textsubscript{char}} & \textbf{Rem\textsubscript{word}} & \textbf{Rem\textsubscript{char}} \\
        BART & 16.6 & 28.4 & 55.6 & 43.4 \\
        RAG wiki & 23.8 & 37.8 & 40.3 & 26.3 \\
        RAG dict & 22.1 & 35.9 & 40.8 & 26.8 \\
        \midrule
        \bottomrule
    \end{tabular}
    \caption{Performance of baseline systems on the Crossword Puzzle dataset. We report the exact-match metric for top-20 predictions of the baseline models listed.}
    \label{tab:crossword_puzzle_baselines}
\end{table}

\begin{table*}[ht]
    \small
    \centering
    \begin{tabular}{lllllllllll}
    \toprule
    Model & Fact. & Hist. & Meaning & Syn./Ant. & Blank & Abbr. & Pref./Suf. & Wordplay & X-lingual & Dependent \\
    \midrule
    BART & 40.4 & 19.0 & 43.9 & 40.3 & 36.0 & 42.9 & 20.0 & 33.5 & 40.0 & 0.0 \\
    RAG-wiki & 53.9 & 28.6 & 55.3 & 46.6 & 60.0 & 60.0 & 60.0 & 43.9 & 60.0 & 11.8\\
    RAG-dict & 54.2 & 35.7 & 52.8 & 48.9 & 61.3 & 85.7 & 60.0 & 46.3 & 40.0 & 11.8 \\
    \bottomrule
    \end{tabular}
    \caption{Performance of models across clue types in the exact match, top-20 setting. Evaluation performed on a 1000 clue subset of the test set which were manually annotated across clue categories. }
    \label{tab:results_by_type}
\end{table*}

\subsection{Crossword Puzzle Task}
The baseline performance on the entire crossword puzzle dataset shows there is significant room for improvement of the existing architectures (see \autoref{tab:crossword_puzzle_baselines}). Our best model, RAG-wiki, correctly fills in the answers for only 26\% (on average) of the total number of puzzle clues, despite having a much higher performance on the clue-answer task, i.e.  measured independently from the crossword grid (\autoref{tab:clue_answering_all_models}). This is explained by the fact that the clues with no ground-truth answer present among the candidates have to be removed from the puzzles in order for the solver to converge, which in turn relaxes the interdependency constraints too much, so that a filled answer may be selected from the set of candidates almost at random. Despite that, the baseline solver is able to solve over a quarter of each the puzzle on average.


\section{Qualitative analysis}
Evaluation on the annotated subset of the data reveals that some clue types present significantly higher levels of difficulty than others (see \autoref{tab:results_by_type}). In particular, all of our baseline systems struggle with the clues requiring reasoning in the context of historical knowledge.
As expected, all of the models demonstrate much stronger performance on the factual and word-meaning clue types, since the relevant answer candidates are likely to be found in the Wikipedia data used for pre-training. 
We observe the biggest differences between BART and RAG performance for the ``abbreviation'' and the ``prefix-suffix'' categories. The document retrieval step in RAG allows for more efficient matching of supporting documents, leading to generation of more relevant answer candidates. For instance, the clue \textit{``Warehouse abbr.''} results in \textit{``pkg''} and \textit{``bldg''} candidates among RAG predictions, whereas BART generates abstract and largely irrelevant strings.

Our manual inspection of model predictions suggest that both BART and RAG correctly infer the grammatical form of the answer from the formulation of the clue. For example, the clue \textit{``Stitched''} produces the candidate answers \textit{``Sewn''} and \textit{``Made''}, and the clue \textit{``Word repeated after ``Que''''} triggers mostly Spanish and French generations (e.g. \textit{``Avec''} or \textit{``Sera''}). 

As previously stated RAG-wiki and RAG-dict largely agree with each other with respect to the ground truth answers. We qualitatively assessed instances where either RAG-wiki or RAG-dict predict the answer correctly in \autoref{sec:appendix}. 

\section{Discussion and Future Work}

The presented task is challenging to approach in an end-to-end model fashion. There are several reasons for this, which we discuss below.

\paragraph{Character-level outputs.} Commonly used Transformer decoders do not produce character-level outputs and produce BPE and wordpieces instead, which creates a problem for a potential end-to-end neural crossword solver.
One possible solution can be the modification of the loss term,
designed with character-based output logits instead of BPE since the crossword grid constraints are at a single cell- (i.e. character-) level. There is some work done in the character-level output transformer encoders such as \citet{ma-etal-2020-charbert}. However, to our best knowledge there is no major generative Transformer architecture which supports character-level outputs yet, we intend to explore this avenue further in future work to develop an end-to-end neural crossword solver.

\paragraph{SMT solver constraints.}

As mentioned earlier, our current baseline solver does not allow partial solutions, and we rely on pre-filtering using the oracle from the ground-truth answers.
%
Although this strategy is flawed for the obvious use of the oracle, the alternatives are currently either computationally intractable or too lossy. One such strategy is to remove $k$ clues at a time, starting with $k=1$ and progressively increasing the number of clues removed until the remaining relaxed puzzle can be solved -- which has the complexity of O($2^n$), where $n$ is the total number of clues in the puzzle. 
Another approach we tried was to relax certain constraints of the puzzle grid, maximally satisfying as many constraints as possible, which is formally known as the maximal satisfaction problem (MAX-SAT). This is a NP-hard problem for which it is hard to find approximate solutions \cite{papadimitriou1994}. 

Our initial foray into such approximate solvers \cite{previti2013partial, liffiton2013enumerating} produced 
severely under-constrained puzzles with garbage character entries.
%
Further work needs to be done to extend this solver to handle partial solutions elegantly without the need for an oracle, this could be addressed with probabilistic and weighted constraint satisfaction solvers, in line with the work by \citet{LITTMAN200223, keim1999cruciverbalist} and \citet{ginsberg2011dr}, but without the dependency on the past crossword clues.

\section{Conclusion}

We present a new challenging task of solving crossword puzzles and present the New York Times Crosswords Dataset, which can be approached at a QA-like level of individual clue-answer pairs, or at the level of an entire puzzle, with imposed answer interdependency constraints. This new benchmark contains a broad range of clue types that require diverse reasoning components. We carry out a set of baseline experiments that indicate the overall difficulty of this task for the current systems, including retrieval-augmented SOTA models for open-domain question answering. 
We also discuss the technical challenges in building a crossword solver and obtaining partial solutions as well as in the design of end-to-end systems for this task. 
We hope that the NYT Crosswords task would define a new high bar for the AI systems.

\section{Ethical Considerations}
The New York Times daily crossword puzzles are a copyright of the New York Times. We have obtained preliminary approval from the New York Times to release this data under a non-commercial and research use license, and are in the process of finalizing the exact licensing terms and distribution channels with the NYT legal department.

\section{Acknowledgments} 
We would like to thank the anonymous reviewers for their careful and insightful review of our manuscript and their feedback. We would like to thank Parth Parikh for the permission to modify and reuse parts of their crossword solver\footref{z3solver}. We are grateful to New York Times staff for their support of this project. This project is funded in part by an NSF CAREER award to Anna Rumshisky (IIS-1652742).

\bibliography{custom}
\bibliographystyle{acl_natbib}

\appendix

\section{Qualitative Analysis of RAG-wiki and RAG-dict Predictions }
\label{sec:appendix}

\begin{table*}[hb]
\centering
\caption{Examples where either RAG-dict or RAG-wiki predicts correctly and other fails.}
\label{tab:examples-for-wiki-n-dict}
\begin{tabular}{l|l|l|l|l} 
\toprule
Category           & \multicolumn{2}{l|}{\begin{tabular}[c]{@{}l@{}}RAG-dict predicts correctly\\RAG-wiki fails\end{tabular}}                                     & \multicolumn{2}{l}{\begin{tabular}[c]{@{}l@{}}RAG-wiki predicts correctly\\RAG-dict fails\end{tabular}}                                         \\ 
\hline
                   & Clue                                                                                 & Answer                                                & Clue                                                                                   & Answer                                                 \\ 
\cline{2-5}
Factual            & \begin{tabular}[c]{@{}l@{}}Asian nursemaid\\Pill alternative, for short\end{tabular} & \begin{tabular}[c]{@{}l@{}}amah\\iud\end{tabular}     & \begin{tabular}[c]{@{}l@{}}Quisling’s city\\Avatar of Vishnu\end{tabular}              & \begin{tabular}[c]{@{}l@{}}oslo\\rama\end{tabular}     \\ 
\hline
Word Meaning       & \begin{tabular}[c]{@{}l@{}}Pause indicator\\Moves along quickly\end{tabular}         & \begin{tabular}[c]{@{}l@{}}comma\\scoots\end{tabular} & \begin{tabular}[c]{@{}l@{}}Sites for grand entrances\\Point of no return?\end{tabular} & \begin{tabular}[c]{@{}l@{}}archways\\ace\end{tabular}  \\ 
\hline
Word Play          & \begin{tabular}[c]{@{}l@{}}Kind of contribution\\Without ice\end{tabular}            & \begin{tabular}[c]{@{}l@{}}ira\\neat\end{tabular}     & \begin{tabular}[c]{@{}l@{}}I’m impressed!\\Airport no no\end{tabular}                  & \begin{tabular}[c]{@{}l@{}}ooh\\knife\end{tabular}     \\ 
\hline
Synonyms Antonyms  & \begin{tabular}[c]{@{}l@{}}Stitched\\Promptly\end{tabular}                           & \begin{tabular}[c]{@{}l@{}}sewn\\on time\end{tabular} & \begin{tabular}[c]{@{}l@{}}guess\\\end{tabular}                                        & \begin{tabular}[c]{@{}l@{}}idea\\\end{tabular}         \\ 
\hline
Fill in the Blanks & \begin{tabular}[c]{@{}l@{}}\_\_rug\\canola \_\_~\end{tabular}                        & \begin{tabular}[c]{@{}l@{}}area\\oil\end{tabular}     & \begin{tabular}[c]{@{}l@{}}\_\_-Israeli relations\\\end{tabular}                       & \begin{tabular}[c]{@{}l@{}}arab\\\end{tabular}         \\
\bottomrule
\end{tabular}
\end{table*}
We examined top-20 exact-match predictions generated by RAG-wiki and RAG-dict. With some exceptions, both models predict similar results (in terms of answer matches) for around 85\% of the test set.

\autoref{tab:examples-for-wiki-n-dict} shows examples where RAG-dict failed to generate the correct predictions but RAG-wiki succeeded, and vice-versa. Most of the instances where RAG-dict predicted correctly and RAG-wiki did not are the ones where answer is closely related to the meaning of the clue. The instances where only RAG-wiki predicted correctly are where answer is not a direct meaning of the clue, and some more information is required predict.

\end{document}